\documentclass[preprint,12pt]{elsarticle}

\usepackage{amssymb}
\usepackage{amsthm}

\usepackage{lineno}

\usepackage{extra}
\usepackage{graphicx}
\usepackage{array}
\usepackage{booktabs}
\usepackage{multirow}
\usepackage{url}
\usepackage{hyperref}
\usepackage{xcolor}
\usepackage{longtable}
\usepackage{tabularx}
\usepackage[nolist,nohyperlinks]{acronym}
\usepackage{amsmath}
\usepackage{subcaption}

\hypersetup{
    colorlinks=true,
    linkcolor=blue,
    filecolor=magenta,      
    urlcolor=magenta,
    citecolor=blue,
    pdfpagemode=FullScreen,
    }

\journal{Data In Brief}

\begin{document}
\begin{frontmatter}


\title{\textbf{MEVDT}: \textbf{M}ulti-Modal \textbf{E}vent-Based \textbf{V}ehicle \textbf{D}etection and \textbf{T}racking Dataset}


\author{Zaid A. El Shair\corref{cor1}\fnref{affil}}
\ead{zelshair@umich.edu}

\author{Samir A. Rawashdeh\fnref{affil}}
\cortext[cor1]{Corresponding author.}

\affiliation[affil]{organization={Department of Electrical and Computer Engineering, University of Michigan-Dearborn},
            addressline={4901 Evergreen Rd}, 
            city={Dearborn},
            postcode={48128}, 
            state={Michigan},
            country={USA}}

\acrodef{2D}{Two-dimensional}
\acrodef{3D}{Three-dimensional}
\acrodef{4D}{Four-dimensional}
\acrodef{AD}{Automated Driving}
\acrodef{ADAS}{Advanced Driver Assistance Systems}
\acrodef{AER}{Address-Event Representation}
\acrodef{AP}{Average Precision}
\acrodef{APS}{Active Pixel Sensor}    
\acrodef{ATIS}{Asynchronous Time-Based Image Sensor}
\acrodef{AVs}{Autonomous Vehicles}
\acrodef{BB}{Bounding Box}
\acrodef{CNN}{Convolutional Neural Network}
\acrodef{CNNs}{Convolutional Neural Networks}
\acrodef{CSV}{Comma-Separated Values}
\acrodef{CV}{Computer Vision}
\acrodef{DAVIS}{Dynamic and Active-Pixel Vision Sensor}
\acrodef{DL}{Deep Learning}
\acrodef{DNN}{Deep Neural Network}
\acrodef{DVS}{Dynamic Vision Sensor}
\acrodef{FN}{False Negative}    
\acrodef{FP}{False Positive}    
\acrodef{FPS}{Frames Per Second}
\acrodef{GCN}{Graph Convolutional Network}
\acrodef{GCNs}{Graph Convolutional Networks}
\acrodef{HDR}{High Dynamic Range}
\acrodef{IoU}{Intersection over Union}
\acrodef{mAP}{mean Average Precision}
\acrodef{ML}{Machine Learning}
\acrodef{MOT}{Multi-Object Tracking}
\acrodef{NMS}{Non-Maximum Suppression}
\acrodef{PNG}{Portable Network Graphics}
\acrodef{ROI}{Region of Interest}
\acrodef{ROS}{Robot Operating System}
\acrodef{SLAM}{Simultaneous Localization and Mapping}
\acrodef{TN}{True Negative}    
\acrodef{TP}{True Positive}

\acrodef{MEVDT}{Multi-Modal Event-based Vehicle Detection and Tracking}

\begin{abstract}

In this data article, we introduce the \ac{MEVDT} dataset. This dataset provides a synchronized stream of event data and grayscale images of traffic scenes, captured using the \ac{DAVIS} 240c hybrid event-based camera. \ac{MEVDT} comprises 63 multi-modal sequences with approximately 13k images, 5M events, 10k object labels, and 85 unique object tracking trajectories. Additionally, \ac{MEVDT} includes manually annotated ground truth labels --- consisting of object classifications, pixel-precise bounding boxes, and unique object IDs --- which are provided at a labeling frequency of 24 Hz. Designed to advance the research in the domain of event-based vision, \ac{MEVDT} aims to address the critical need for high-quality, real-world annotated datasets that enable the development and evaluation of object detection and tracking algorithms in automotive environments.

\end{abstract}




\end{frontmatter}





\newpage\noindent\textbf{Specifications Table}

\small{
\noindent\begin{tabular}[t]{p{0.28\textwidth} p{0.72\textwidth}}
\toprule
 Subject & Computer Vision and Pattern Recognition, Computer Science Applications,
 Signal Processing,        
 Artificial Intelligence
 \smallskip \\

 Specific subject area & Event-Based and Multi-Modal Object Detection and Tracking.\\

 Type of data & \acs{2D}-Grayscale Images (.png), Event Streams (.csv), Fixed-duration Event Files (.aedat), Sequence Annotations (.txt), Sample Annotations (.txt), Train-Test Split Files (.csv) \smallskip \\
 
 Data collection
& 

Data was collected using the hybrid sensor \ac{DAVIS} 240c, which combines an \ac{APS} and a \ac{DVS} within the same pixel array. The \ac{APS} captures grayscale images at 24 FPS, while the \ac{DVS} records pixel brightness changes (\ie, events) at microsecond resolution. The collection process took place at the University of Michigan-Dearborn campus, in two scenes under clear daylight conditions. Data recording was managed using the \ac{ROS} \ac{DVS} package running on a laptop. The camera was fixed capturing moving vehicles which were manually labeled with 2D bounding boxes and unique IDs. \smallskip \\

 Data source location & Country: United States of America \smallskip \\
 
 Data accessibility
& 
Repository name: Deep Blue Data

Data identification number: 10.7302/d5k3-9150

Direct URL to data: \href{https://doi.org/10.7302/d5k3-9150}{https://doi.org/10.7302/d5k3-9150}

 \smallskip \\

 Related research article
& Z. El Shair, S. A. Rawashdeh, High-Temporal-Resolution Object Detection and Tracking Using Images and Events, Journal of Imaging 8 (8) (2022): 210 \cite{el2022high}\\
\bottomrule
\end{tabular}
\label{tab:specifications}
}


\section{Value of the Data}
\begin{itemize}
    \item The \ac{MEVDT} dataset facilitates the development of models specifically tailored for event-based vision, a cutting-edge imaging technology inspired by the human retina \cite{EventSurveygallego2020}. This dataset provides high temporal resolution and asynchronous event data, essential for capturing dynamic changes in a scene, thereby advancing research in event-based \ac{CV} tasks such as object detection and tracking.

    \item Event-based vision is a novel visual sensing modality that offers distinct advantages over conventional frame-based modality, including high dynamic range and robustness to motion blur \cite{EventSurveygallego2020}. \ac{MEVDT} is one of the few event-based datasets for object detection \cite{de2020large, perot2020learning, li2023sodformer} and one of the very few multi-object tracking datasets publicly available \cite{mueggler2017event, gao2023reconfigurable}.

    \item \ac{MEVDT} provides comprehensive annotations for object tracking. Unlike many existing datasets \cite{mueggler2017event, binas2017ddd17, sironi2018hats}, \ac{MEVDT} includes detailed annotations with 2D bounding boxes and unique object IDs. These comprehensive labels are crucial for developing and evaluating object-tracking algorithms, making this dataset a valuable resource for researchers working on high-speed perception and tracking applications.

    \item By combining asynchronous events from a \ac{DVS} with synchronous grayscale frames from an \ac{APS}, the \ac{MEVDT} dataset supports research into multi-modal data fusion. This capability is important for enhancing the accuracy and robustness of computer vision systems, particularly in challenging conditions where traditional vision systems may struggle.
    
    \item Researchers and practitioners in the domain of event-based vision can utilize this dataset to develop event-based and multi-modal solutions for object detection and tracking. Additionally, the provided test set can be used to evaluate their model's performance.

\end{itemize}



\section{Background}

Event-based vision represents a paradigm shift in visual sensing technology, where sensors inspired by the human retina capture dynamic changes in a scene at high temporal resolutions \cite{EventSurveygallego2020, DVS128-lichtsteiner2008128}. Unlike traditional frame-based cameras, event-based sensors detect per-pixel brightness changes, offering advantages in dynamic range and temporal resolution \cite{EventSurveygallego2020}. This emerging field necessitates specialized datasets to promote research and development, particularly in the \ac{CV} tasks of object detection and tracking.

Object detection and tracking are critical in various applications, including \ac{AD} and traffic monitoring\cite{el2022high, marti2019review}. Event cameras, due to their low latency and high-temporal-resolution output, offer promising prospects in these areas. However, the development of relevant methodologies has been impeded by a lack of labeled event-based datasets tailored to these applications. Existing datasets often lack the necessary annotations, such as object IDs, essential for tracking applications \cite{de2020large, perot2020learning, li2023sodformer, binas2017ddd17, sironi2018hats, miao2019neuromorphic}.

To address this limitation, we have created a dataset specifically tailored for event-based and multi-modal object detection and tracking. \ac{MEVDT} includes many sequences with multiple vehicles moving at various speeds while featuring manually labeled bounding boxes and object IDs, which are vital for enabling object tracking evaluation.


In contrast to our prior publications \cite{el2022high, el2023high}, which presented methods that utilize this dataset, this article provides a comprehensive overview and in-depth analysis of the \ac{MEVDT} dataset itself. This article offers a detailed breakdown of the dataset's statistics and introduces a sequence-based training/testing split, facilitating its use in model development and evaluation.

\section{Data Description}\label{sec:Data_Description}


This section outlines the \ac{MEVDT} dataset's structure, offering detailed insights into its organization and content to support various \ac{CV} tasks. First, we describe the main statistics of the dataset and detail each sequence's characteristics (Section \ref{sec:dataset_statistics}). Then, we present the dataset's directory structure by outlining each subdirectory's purpose and contents (Section \ref{sec:dataset_structure}). Finally, we detail the different data sample formats (Section \ref{sec:data_format}) and label formats (Section \ref{sec:label_format}) available in \ac{MEVDT}.


\subsection{Dataset Statistics}\label{sec:dataset_statistics}

The \ac{MEVDT} dataset contains multiple recordings of numerous vehicles moving at varying speeds captured at two different scenes. These recordings are segmented into shorter sequences for a more focused analysis and usage. Accordingly, Scene A includes 32 sequences, comprising 9,274 images and 6,828 annotations, while Scene B contains 31 sequences with 3,485 images and 3,063 annotations. Additionally, each sequence includes continuous streams of asynchronous events captured by the hybrid camera. Overall, our dataset provides a total of 9,891 vehicle annotations. The discrepancy between the number of images and annotations arises from frames without detectable objects.

\begin{table}[b]
\centering
\caption{Sequence statistics for Scenes A and B in the \ac{MEVDT} dataset. This table details the total number of sequences for each scene and provides the total and average (with standard deviations) sequence duration, number of images, events, objects, object IDs, and average bounding box area.}
\label{tab:Dataset:statistics}
\resizebox{\textwidth}{!}{
    \begin{tabular}{l >{\centering\arraybackslash}m{12 mm}  c  c  c c  c c c c  c c>{\centering\arraybackslash}m{26 mm}}
    \toprule
    \multirow{2}{*}{\textbf{Subset}} & \multirow{2}{12 mm}{\centering\textbf{\# of Seqs.}} & \multicolumn{2}{c}{\textbf{Seq. Duration $(s)$}} & \multicolumn{2}{c}{\textbf{\# of Images}} & \multicolumn{2}{c}{\textbf{\# of Events}} & \multicolumn{2}{c}{\textbf{\# of Objects}} & \multicolumn{2}{c}{\textbf{\# of Object IDs}}&\multirow{2}{26 mm}{\centering\textbf{Average BB Area $(pixel^2)$}} \\
    \cmidrule(lr){3-4}
    \cmidrule(lr){5-6}
    \cmidrule(lr){7-8}
    \cmidrule(lr){9-10}
    \cmidrule(lr){11-12}
     &  & Total & Average $\pm$ SD & Total & Average $\pm$ SD & Total & Average $\pm$ SD & Total & Average $\pm$ SD &   Total &Average $\pm$ SD &\\
     \midrule
    Scene A & 32 & 397.3 & 12.42 $\pm$9.94 & 9274 & 289.81 $\pm$230.8 & 2269913 & \ \ 70935 $\pm$59337 & 6828 & 213 $\pm$147 &   54
&1.7 $\pm$1.1
&1960.5 \\
    Scene B & 31 & 147.7 & \ \ 4.76 $\pm$3.55 & 3485 & 112.42 $\pm$82.4 \ \ & 3195652 & 103086 $\pm$31950 & 3063 & 99 $\pm$84 &   31
&1 $\pm$0
&4093.2 \\
    \cmidrule(lr){1-13}
    \textbf{Total} & 63 & 545.0 & \ \ 8.65 $\pm$8.39 & 12759 & 202.52 $\pm$194.7 & 5465565 & \ \ 86755 $\pm$50169
 & 9891 & 157 $\pm$132 &   85
&1.3 $\pm$0.8
&3010.0 \\
    \bottomrule
    \end{tabular}
}
\end{table}

\begin{table}[t]
\centering
\caption
{Training and testing split statistics for \textbf{Scene A}. This table provides a comprehensive breakdown of the sequences in this location, including their durations, number of images, events, objects, object IDs, and average bounding box areas for both training and testing subsets.}
\label{tab:Dataset:SceneA_train_test}
\resizebox{\textwidth}{!}{
\begin{tabular}{l c c c c c >{\centering\arraybackslash}m{24 mm} >{\centering\arraybackslash}m{26 mm}}
\toprule
 & \textbf{\# of Seqs.} & \textbf{Seq. Duration $(s)$}& \textbf{\# of Images}& \textbf{\# of Events}&\textbf{\# of Objects}&  \textbf{\# of Object IDs}&\textbf{Average BB Area $(pixel^2)$}\\

\midrule
 \multicolumn{8}{c}{\textbf{Training}} \\
\cmidrule(lr){1-8}
Average & $-$ & 12.2 & 284.0 & 72512.3 & 210.8 &  1.7 &1932.7 \\
SD & $-$ & 9.8 & 227.4 & 63088.0 & 138.8 &  1.0 &914.8 
\\
Total &  26 & 316.4 & 7385& 1885319& 5481&  44&50249.8 
\\
\% & 81\% & 80\% & 80\% & 83\% & 80\% &  81\% &80\% \\


\midrule
 \multicolumn{8}{c}{\textbf{Testing}} \\
\cmidrule(lr){1-8}
Average & $-$ & 13.5 & 314.8 & 64099.0 & 224.5 &  1.7
&2081.1 \\
SD & $-$ & 9.6 & 224.1 & 29780.6 & 166.1 &  1.1
&1109.4 \\
Total & 6 & 80.9 & 1889 & 384594 & 1347 &  10
&12486.5 \\
\% & 19\% & 20\% & 20\% & 17\% & 20\% &  19\%&20\% \\
\bottomrule
\end{tabular}
}
\end{table}

\begin{table}[!htb]
\centering
\caption{Training and testing split statistics for \textbf{Scene B}. This table provides a comprehensive breakdown of the sequences in this location, including their durations, number of images, events, objects, object IDs, and average bounding box areas for both training and testing subsets.}
\label{tab:Dataset:SceneB_train_test}
\resizebox{\textwidth}{!}{
\begin{tabular}{l c c c c c >{\centering\arraybackslash}m{24 mm} >{\centering\arraybackslash}m{26 mm}}
\toprule
 & \textbf{\# of Seqs.} & \textbf{Seq. Duration $(s)$}& \textbf{\# of Images}& \textbf{\# of Events}&\textbf{\# of Objects}&  \textbf{\# of Object IDs}&\textbf{Average BB Area $(pixel^2)$}\\

\midrule
 \multicolumn{8}{c}{\textbf{Training}} \\
\cmidrule(lr){1-8}
Average & $-$  & 4.7 & 109.9 & 103113.4 & 96.0 &  1 &4088.3 
\\
SD & $-$ & 3.3 & 76.4 & 32434.7 & 77.1 &  0 &1123.9 
\\
Total & 25& 116.3 & 2747& 2577836& 2400&  25 &102208.1 
\\
\% & 81\% & 79\% & 79\% & 81\% & 78\% &  81\% &81\% \\
\midrule

 \multicolumn{8}{c}{\textbf{Testing}} \\
\cmidrule(lr){1-8}
Average & $-$  & 5.2 & 123.0 & 102969.3 & 110.5 &  1 &4113.5 
\\
SD  & $-$ & 4.2 & 97.5 & 26843.0 & 99.9 &  0 &1011.2 
\\
Total & 6 & 31.3 & 738 & 617816 & 663 &  6 &24681.1 
\\
\% & 19\% & 21\% & 21\% & 19\% & 22\% &  19\% &19\% \\

\bottomrule
\end{tabular}
}
\end{table}

The sequence statistics for each scene, including sequence durations, number of images, events, and objects, are summarized in Table \ref{tab:Dataset:statistics}. On average, each generated sequence is approximately 9 seconds in length with around 200 images and 87,000 events. This translates to an average event rate of 10,000 events per second, underscoring the high temporal resolution characteristic of event-based sensors. As a result of our labeling, the dataset provides 85 different continuous object trajectories in total, each represented by a unique object ID.

Additionally, we provide sequence-based training and test splits. These splits are critical for methodical model development and evaluation, ensuring reproducibility and consistency in experimental setups. We allocate approximately 80\% of the sequences for training and 20\% for testing in both Scene A and Scene B, ensuring a well-balanced distribution of images, events, and objects. This balance is achieved by establishing an appropriate 80\%--20\% (4:1) distribution across all of the dataset's parameters. Details of this distributation are demonstrated in Table \ref{tab:Dataset:SceneA_train_test} and Table \ref{tab:Dataset:SceneB_train_test} for Scene A and Scene B, respectively. The 80--20 split, a standard heuristic in machine learning, strikes a balance between the need for ample training data and representative testing data, promoting robust model training, preventing overfitting, and enabling reliable model validation and generalization to unseen data.


Detailed breakdown of each sequence in Scene A and Scene B are demonstrated in Table \ref{tab:Dataset:sceneA} and Table \ref{tab:Dataset:sceneB}, respectively. These breakdowns provide detailed information on each sequence, including the sequence name (identified by the first data timestamp in nanoseconds), duration, number of images, events, objects, and the average area of bounding boxes, along with their allocation to either training or testing splits. This detailed information aids in understanding the dataset composition and its distribution between training and testing.





\begin{table}[t!]
\centering
\caption{Detailed sequence statistics for \textbf{Scene A}. This table includes information on each sequence's duration, number of images, events, objects, object IDs, average bounding box area, and allocation to training or testing splits.} \label{tab:Dataset:sceneA}
\resizebox{\textwidth}{!}{
\begin{tabular}{>{\centering\arraybackslash}m{25 mm} >{\centering\arraybackslash}m{37 mm} >{\centering\arraybackslash}m{26 mm} >{\centering\arraybackslash}m{15 mm} >{\centering\arraybackslash}m{15 mm} >{\centering\arraybackslash}m{17 mm} >{\centering\arraybackslash}m{23 mm} >{\centering\arraybackslash}m{26 mm} >{\centering\arraybackslash}m{24 mm}}
\toprule
\textbf{Sequence \#} & \textbf{Sequence Name} & \textbf{Duration ($s$)} & \textbf{\# of Images} & \textbf{\# of Events} & \textbf{\# of Objects} &  \textbf{\# of Object  IDs} &\textbf{Average BB Area ($pixel^2$)} & \textbf{Train $|$ Test} \\
\midrule
1 & 1581956305832790936 & 9.5 & 222 & 76924 & 240 &  2
&1852.4 & Train \\
2 & 1581956366514475936 & 21.2 & 494 & 48556 & 122 &  1
&1562.3 & Train \\
3 & 1581956422501835936 & 10.4 & 243 & 70138 & 241 &  2
&1490.3 & Test \\
4 & 1581956475991297936 & 23.3 & 542 & 61664 & 135 &  1
&1754.2 & Test \\
5 & 1581956525690846936 & 17.4 & 404 & 79738 & 382 &  2
&1476.3 & Train \\
6 & 1581956568112038936 & 5.3 & 124 & 34207 & 117 &  1
&1394.1 & Test \\
7 & 1581956586329463936 & 12.1 & 283 & 114163 & 186 &  1
&2997.0 & Train \\
8 & 1581956636804222936 & 4.9 & 115 & 29212 & 102 &  1
&1522.7 & Train \\
9 & 1581956672808401936 & 5.4 & 127 & 60633 & 118 &  1
&1665.5 & Train \\
10 & 1581957068983574936 & 21.0 & 488 & 154064 & 420 &  3
&3373.5 & Train \\
11 & 1581957114204134936 & 7.0 & 163 & 34310 & 160 &  1
&1609.8 & Train \\
12 & 1581957156969863936 & 8.3 & 195 & 62531 & 192 &  1
&4538.2 & Test \\
13 & 1581957173378467936 & 2.5 & 59 & 20295 & 58 &  1
&1315.7 & Train \\
14 & 1581957190648414936 & 45.6 & 1061 & 107768 & 224 &  1
&1796.1 & Train \\
15 & 1581957249133671936 & 6.7 & 158 & 51306 & 150 &  1
&1730.8 & Train \\
16 & 1581957506675527936 & 18.1 & 421 & 77074 & 502 &  3
&1373.6 & Train \\
17 & 1581957567314145936 & 4.1 & 96 & 34093 & 81 &  1
&1710.7 & Train \\
18 & 1581957616841425936 & 10.3 & 241 & 41452 & 208 &  2
&1386.9 & Train \\
19 & 1581957903798179936 & 6.2 & 145 & 31237 & 130 &  1
&1658.5 & Train \\
20 & 1581957963058646936 & 5.2 & 124 & 48102 & 122 &  1
&2071.8 & Train \\
21 & 1581958023266591936 & 4.6 & 109 & 20877 & 97 &  1
&1476.4 & Train \\
22 & 1581958094284404936 & 5.1 & 119 & 14818 & 113 &  1
&1587.9 & Train \\
23 & 1581958106816959936 & 14.9 & 348 & 140138 & 399 &  3
&1396.0 & Train \\
24 & 1581958201263329936 & 25.4 & 592 & 91349 & 239 &  3
&1737.3 & Train \\
25 & 1581958201392531936 & 2.8 & 65 & 23145 & 64 &  1
&1608.6 & Train \\
26 & 1581958289206551936 & 14.9 & 346 & 71577 & 282 &  3
&1816.2 & Train \\
27 & 1581958320817876936 & 5.4 & 126 & 76807 & 113 &  1
&5773.9 & Train \\
28 & 1581958380465948936 & 29.7 & 694 & 122720 & 578 &  4
&1487.0 & Test \\
29 & 1581958511820908936 & 9.7 & 226 & 78420 & 198 &  2
&1541.7 & Train \\
30 & 1581958540632865936 & 3.9 & 91 & 33334 & 84 &  1
&1822.8 & Test \\
31 & 1581958551959539936 & 29.0 & 676 & 330350 & 605 &  5
&2782.1 & Train \\
32 & 1581958587877583936 & 7.6 & 177 & 28911 & 166 &  1
&1426.3 & Train \\
\bottomrule
\end{tabular}
}
\end{table}

\begin{table}[t!]
\centering
\caption{Detailed sequence statistics for \textbf{Scene B}. This table includes information on each sequence's duration, number of images, events, objects, object IDs, average bounding box area, and allocation to training or testing splits}\label{tab:Dataset:sceneB}

\resizebox{\textwidth}{!}{
\begin{tabular}{>{\centering\arraybackslash}m{25 mm} >{\centering\arraybackslash}m{37 mm} >{\centering\arraybackslash}m{26 mm} >{\centering\arraybackslash}m{15 mm} >{\centering\arraybackslash}m{15 mm} >{\centering\arraybackslash}m{17 mm} >{\centering\arraybackslash}m{23 mm} >{\centering\arraybackslash}m{26 mm} >{\centering\arraybackslash}m{24 mm}} 
\toprule
\textbf{Sequence \#} & \textbf{Sequence Name} & \textbf{Duration ($s$)} & \textbf{\# of Images} & \textbf{\# of Events} & \textbf{\# of Objects} &  \textbf{\# of Object  IDs} &\textbf{Average BB Area ($pixel^2$)} & \textbf{Train $|$ Test} \\
\midrule
1 & 1603470885671858364 & 14.1 & 329 & 130227 & 321 &  1
&5706.2 & Test \\
2 & 1603470907722265364 & 4.9 & 116 & 109775 & 107 &  1
&4216.0 & Train \\
3 & 1603470947042618364 & 8.3 & 195 & 106234 & 188 &  1
&2468.5 & Train \\
4 & 1603471304371177364 & 5.8 & 137 & 123331 & 90 &  1
&3988.2 & Train \\
5 & 1603471325344903364 & 2.1 & 49 & 87050 & 44 &  1
&3017.5 & Train \\
6 & 1603471347223041364 & 2.7 & 65 & 106671 & 55 &  1
&3995.4 & Train \\
7 & 1603471362511897364 & 2.4 & 58 & 91918 & 47 &  1
&3043.1 & Train \\
8 & 1603471387318604364 & 2.5 & 61 & 108368 & 52 &  1
&3971.9 & Train \\
9 & 1603471400411033364 & 2.1 & 51 & 64688 & 34 &  1
&2924.9 & Test \\
10 & 1603471419705138364 & 6.0 & 142 & 91522 & 127 &  1
&5007.8 & Train \\
11 & 1603471437405757364 & 1.7 & 41 & 55782 & 27 &  1
&3028.5 & Train \\
12 & 1603471457905745364 & 6.8 & 159 & 92745 & 142 &  1
&4610.4 & Train \\
13 & 1603471475606364364 & 2.4 & 56 & 67010 & 40 &  1
&3087.3 & Train \\
14 & 1603471489904674364 & 2.9 & 68 & 151037 & 52 &  1
&4502.9 & Train \\
15 & 1603471504116850364 & 6.9 & 163 & 94593 & 152 &  1
&5330.4 & Train \\
16 & 1603471523712426364 & 2.3 & 56 & 67044 & 45 &  1
&2963.8 & Test \\
17 & 1603471544513884364 & 17.6 & 410 & 104188 & 400 &  1
&5260.7 & Train \\
18 & 1603471574445588364 & 2.3 & 56 & 64461 & 43 &  1
&3054.5 & Train \\
19 & 1603471594816373364 & 6.8 & 159 & 105311 & 150 &  1
&5210.7 & Train \\
20 & 1603471817884627727 & 2.8 & 67 & 83430 & 53 &  1
&4643.3 & Train \\
21 & 1603471844801627727 & 3.8 & 91 & 116064 & 76 &  1
&4082.0 & Test \\
22 & 1603471863492792727 & 4.1 & 96 & 216927 & 76 &  1
&7410.9 & Train \\
23 & 1603471880891941727 & 3.6 & 86 & 103624 & 64 &  1
&3000.1 & Train \\
24 & 1603471908885621727 & 6.4 & 151 & 112285 & 142 &  1
&5159.1 & Train \\
25 & 1603471928136659727 & 2.5 & 61 & 72175 & 49 &  1
&3057.4 & Train \\
26 & 1603471965045249727 & 6.8 & 161 & 119105 & 151 &  1
&5128.6 & Train \\
27 & 1603471985975909727 & 3.1 & 73 & 87510 & 47 &  1
&3042.6 & Train \\
28 & 1603472011687027727 & 2.7 & 65 & 113585 & 51 &  1
&3972.8 & Test \\
29 & 1603472029387646727 & 2.8 & 68 & 74378 & 57 &  1
&3059.0 & Train \\
30 & 1603472052643934727 & 6.2 & 146 & 126208 & 136 &  1
&5031.5 & Test \\
31 & 1603472118493683727 & 2.0 & 49 & 148406 & 45 &  1
&3913.5 & Train \\
\bottomrule
\end{tabular}}
\end{table}

\subsection{Dataset Structure}\label{sec:dataset_structure}
The \ac{MEVDT} dataset is structured into several primary directories, each tailored to specific data handling needs:
\begin{itemize}
    \item \textbf{\texttt{sequences/}}: Stores sequences of images and corresponding event streams, serving as the primary data source for object-tracking solutions.
    
    \item \textbf{\texttt{labels/}}: Contains detailed ground truth labels for object detection and object tracking, essential for model training and evaluation.
    
    \item \textbf{\texttt{event\_samples/}}: Features fixed-duration event-data samples extracted at different batch-sampling durations, providing varied temporal resolutions for advanced event-based methodologies.
    
    \item \textbf{\texttt{data\_splits/}}: Includes \ac{CSV} files that index the path of each sample, including the extracted event sample file (provided in \texttt{event\_samples/}), corresponding image (available in \texttt{sequences/}), and detection label file (within \texttt{labels/detection\_labels/}) facilitating straightforward data loading and partitioning.
\end{itemize}

\begin{figure}
    \centering
    \frame{\includegraphics[width=\textwidth]{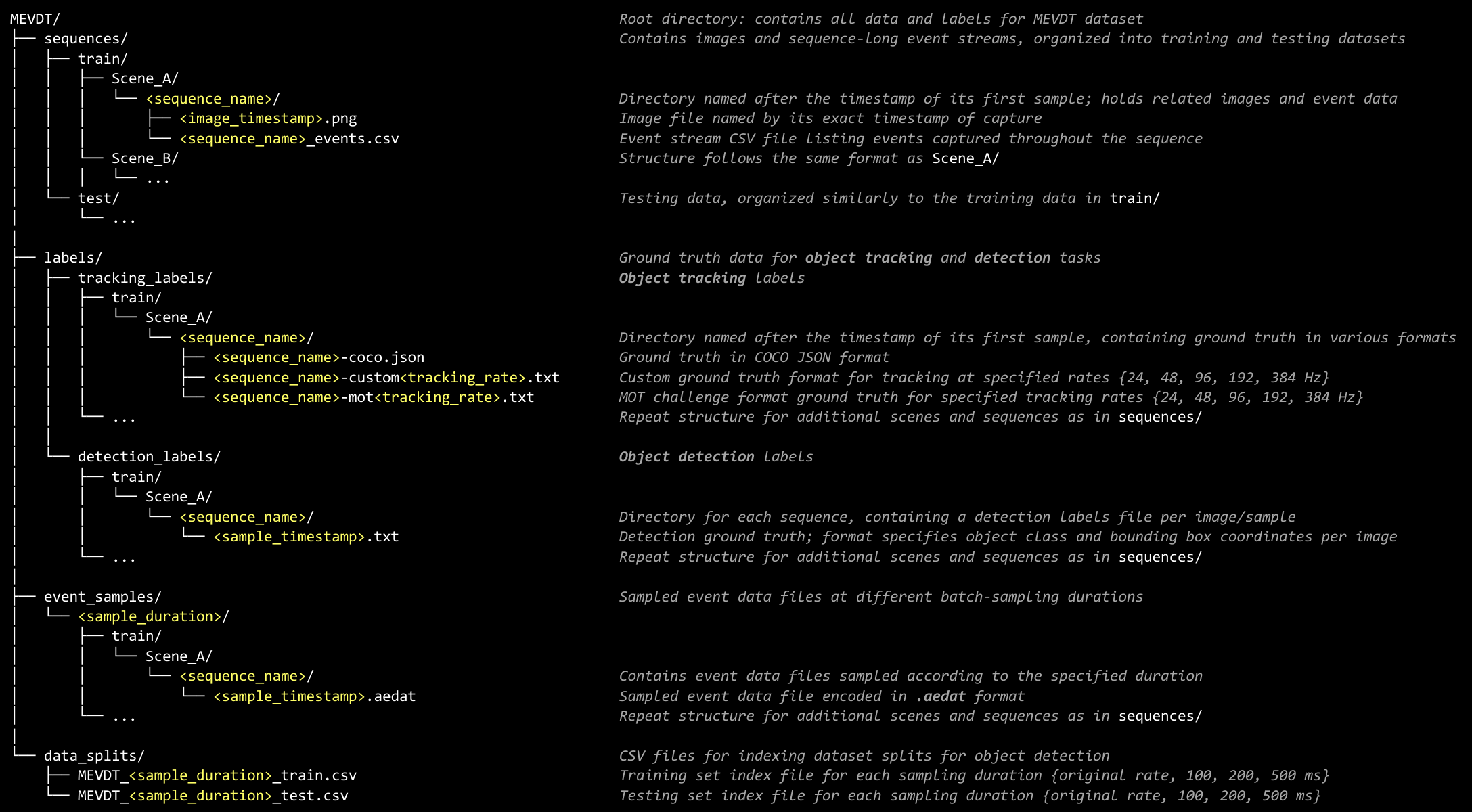}}
    \caption{MEVDT dataset directory structure and organization.}
    \label{fig:dataset_structure}
\end{figure}

Generally, the dataset is split into training and testing splits. Within each split, a list of sequences per the locations \texttt{Scene\_A} and \texttt{Scene\_B} is provided. The full dataset structure and directory breakdown are detailed in Figure \ref{fig:dataset_structure}.

\subsection{Sample Formats}\label{sec:data_format}
The dataset samples are provided in various formats. As shown in Figure \ref{fig:dataset_structure}, the grayscale images are stored in a standard \ac{PNG} format. Meanwhile, the sequence-long event streams are provided in a \ac{CSV} file where each line corresponds to a single event in a comma-separated format as follows:
\begin{equation}
    \texttt{ts,x,y,p},
\end{equation}
where \texttt{ts} denotes the timestamp of the event in nanoseconds, while \texttt{x} and \texttt{y} correspond to the \ac{2D} pixel coordinate at which the event has occurred. Finally, \texttt{p} denotes the polarity of the event as either positive $(\texttt{p} = 1)$ or negative $(\texttt{p}=0)$.

In contrast, the extracted event samples, provided in the \texttt{event\_samples/} directory, are stored in a different format. These samples, generated at various fixed sampling durations, are encoded in \textit{AEDAT} 3.1 file format following the \textit{DVS-Gesture} dataset's implementation \cite{DVS-Gestureamir2017low}.

\subsection{Label Formats}\label{sec:label_format}
The \ac{MEVDT} dataset includes separate ground truth label files for both object detection and object tracking. Each label format is provided in their respective directories within \texttt{labels/}.

\subsubsection{Object Tracking}
As shown in Figure \ref{fig:dataset_structure}, the object tracking label files are available in three formats:
\begin{itemize}
    \item COCO annotations in JSON file format \cite{lin2014microsoft}.
    \item MOT Challenge format \cite{dendorfer2020motchallenge}.
    \item Custom format.
\end{itemize}

The COCO format is well-defined in \cite{lin2014microsoft}. The MOT Challenge format, detailed in \cite{dendorfer2020motchallenge}, is necessary for using \textit{TrackEval} \cite{luiten2020trackeval}, which facilitates the generation of \ac{MOT} results as demonstrated in \cite{el2022high, el2023high}. Our custom format, which utilizes a single line per sample, is defined as follows:

\begin{equation}
    \texttt{frame\_id}, \texttt{frame\_ts}, \{\texttt{obj\_id}_i, \texttt{x}_i, \texttt{y}_i, \texttt{w}_i, \texttt{h}_i\}_{i=1}^n \ ,
\end{equation}
where $\texttt{frame\_id}$ is the frame's unique identifier within the sequence and $\texttt{frame\_ts}$ is the frame's timestamp in nanoseconds. Each set, denoted by $\{\texttt{obj\_id}_i, \texttt{x}_i, \texttt{y}_i, \texttt{w}_i, \texttt{h}_i\}$, specifies the bounding box for the $i^\text{th}$ object, detailing its identifier $\texttt{obj\_id}_i$, top-left corner coordinates $(\texttt{x}_i, \texttt{y}_i)$, width $\texttt{w}_i$, and height $\texttt{h}_i$ when present in the given frame. This format supports an arbitrary number of objects per frame and uses nanosecond timestamps to ensure synchronization with high-temporal-resolution event data. We also provide ground truth for multiple tracking rates (24, 48, 96, 192, and 384 Hz) in both our custom and MOT Challenge \cite{dendorfer2020motchallenge} formats.

\subsubsection{Object Detection}
Object detection labels, as illustrated in Figure \ref{fig:dataset_structure}, consist of a label file per sample. The label format for each object detected in a sample is represented as follows:
\begin{equation}
    \texttt{class\_idx} \ \texttt{x\_min} \ \texttt{y\_min} \ \texttt{x\_max} \ \texttt{y\_max} \ ,
\end{equation}
where $\texttt{class\_idx}$ denotes the object's category (set as \texttt{1} for the vehicle class). The coordinates ($\texttt{x\_min}, \texttt{y\_min}$) and ($\texttt{x\_max}, \texttt{y\_max}$) define the top-left and bottom-right corners of the object's bounding box, respectively. Each line within a file represents a separate object; therefore, the label files may contain multiple lines --- or none --- each corresponding to a single object instance. For example, if a vehicle is labeled within an image with its bounding box starting at (\texttt{50, 50}) and extending to (\texttt{200, 150}), then the label file would contain the following:
\begin{equation}
    \texttt{1 50 50 200 150}
\end{equation}
This line indicates a single vehicle, categorized under index \texttt{1}, occupying the specified pixel range. If multiple vehicles are detected, each will be represented by a similar line within the same file detailing its classification and location. Note that $\texttt{class\_idx}$ skips the index \texttt{0} as it typically represents the \texttt{`background`} class in most object detectors.

\section{Experimental Design, Materials and Methods}
In this section, we detail the data collection and labeling process of the \ac{MEVDT} dataset. We begin with the data collection setup (Section \ref{sec:data_collection_setup}), describing the sensors, settings, and locations used. Following this, we outline the labeling process employed in this work (Section \ref{sec:data_labeling}).

\begin{figure}[t]
    \centering
    \frame{\includegraphics[width=0.75\textwidth]{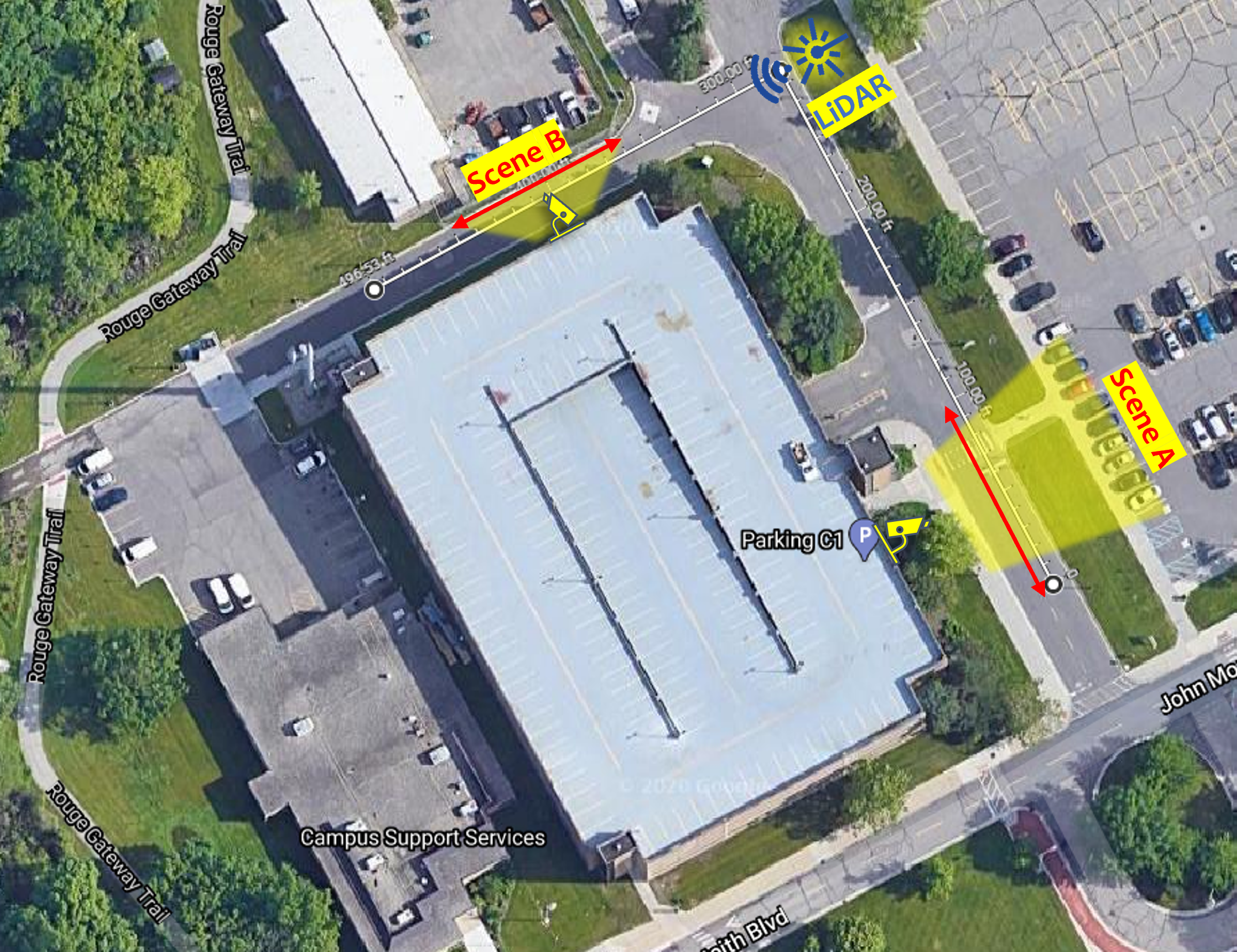}}
    \caption[Satellite view of the data collection locations.]{Satellite view of a subsection of the University of Michigan-Dearborn campus highlighting Scene A and Scene B, where data was collected, along with the position of the LiDAR sensor.}
    \label{fig:dataset: map}
\end{figure}

\subsection{Data Collection Setup}\label{sec:data_collection_setup}


We utilized the hybrid sensor \ac{DAVIS} 240c\footnote{DAVIS 240c specifications available at \href{https://inivation.com/wp-content/uploads/2019/08/DAVIS240.pdf}{https://inivation.com/wp-content/uploads/2019/08/DAVIS240.pdf}}, which combines an \ac{APS} and a \ac{DVS} within the same pixel array. This sensor captures both synchronous intensity frames and asynchronous events, providing a comprehensive visual dataset crucial for developing event-based and multi-modal solutions. The \ac{APS} captures intensity (\ie, grayscale) frames at approximately 24 \ac{FPS}, while the \ac{DVS} records changes in pixel intensity --- known as events --- at a microsecond resolution, essential for capturing changes in the scene at very high speeds.


Additionally, during a subset of the data collection process, we employed a high-speed LiDAR, Benewake TF03-100, to provide high-temporal-resolution ground truth positional measurements. This LiDAR, capable of delivering measurements at rates up to 1000 Hz, was placed 30 to 60 meters from the vehicles and used to precisely estimate distances to the tracked vehicles at high tracking rates. Although these positional measurements offer valuable insights for validating object tracking performance, they are not included in the \ac{MEVDT} dataset as it is aimed primarily at \ac{CV} tasks of object detection and tracking. The primary reason for their omission is to maintain the dataset's focus on visual data processing challenges, as the LiDAR data falls outside the typical usage scenarios intended for users of this dataset. Detailed insights and applications of this LiDAR data in evaluating tracking accuracy are further explored in our prior work \cite{el2023high}.

\begin{figure}[t]
    \centering
     \begin{subfigure}[b]{0.45\textwidth}
         \centering
         \frame{\includegraphics[width=\textwidth]{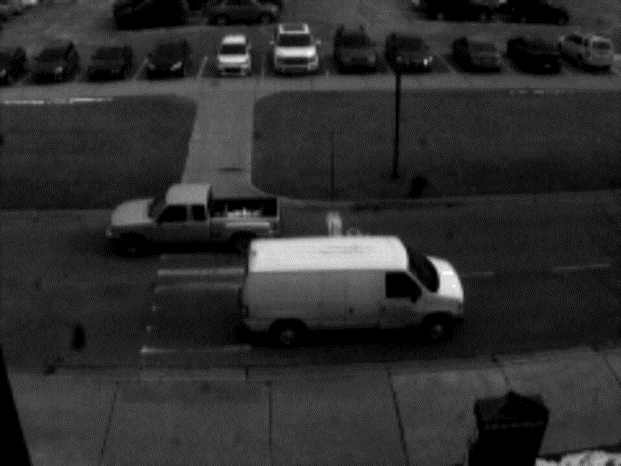}}
         \caption{Scene A}
         \label{fig:Scene_A}
     \end{subfigure}
     \begin{subfigure}[b]{0.45\textwidth}
         \centering
         \frame{\includegraphics[width=\textwidth]{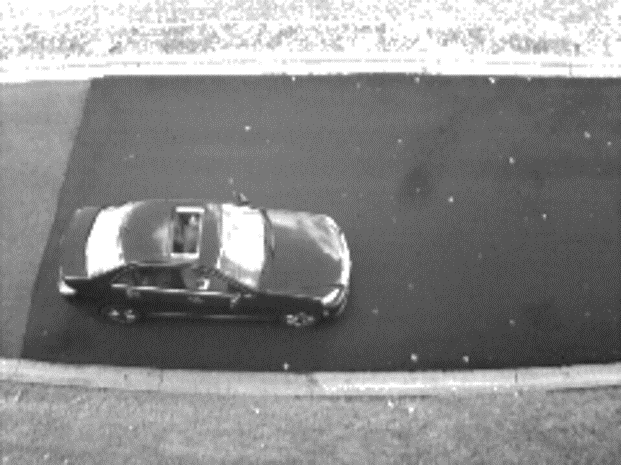}}
         \caption{Scene B}
     \end{subfigure}
    \caption{Sample image outputs from the dataset demonstrating the two distinct scenes, including (a) Scene A and (b) Scene B, showcasing the camera's perspective and field of view for each location within the University of Michigan-Dearborn's campus.}
    \label{fig:Dataset:scenes}
\end{figure}

Using \ac{DAVIS} 240c and the \ac{ROS} \ac{DVS} package developed by Robotics and Perception Group\footnote{Available at \href{https://github.com/uzh-rpg/rpg_dvs_ros}{https://github.com/uzh-rpg/rpg\_dvs\_ros}} \cite{mueggler2014event} to record the data, we collected several hours of spatiotemporally synchronized images and events. The data collection was conducted at two different places within the same location (at the campus of the University of Michigan-Dearborn), referred to as Scene A and Scene B. Each scene was recorded on a different day with generally clear daylight conditions. A satellite map view depicting the data collection location including the positions of each scene is shown in Figure \ref{fig:dataset: map}. Furthermore, we show an image sample from each scene in Figure \ref{fig:Dataset:scenes}.

\begin{figure}[!htb]
    \centering
    \frame{\includegraphics[width=0.65\textwidth]{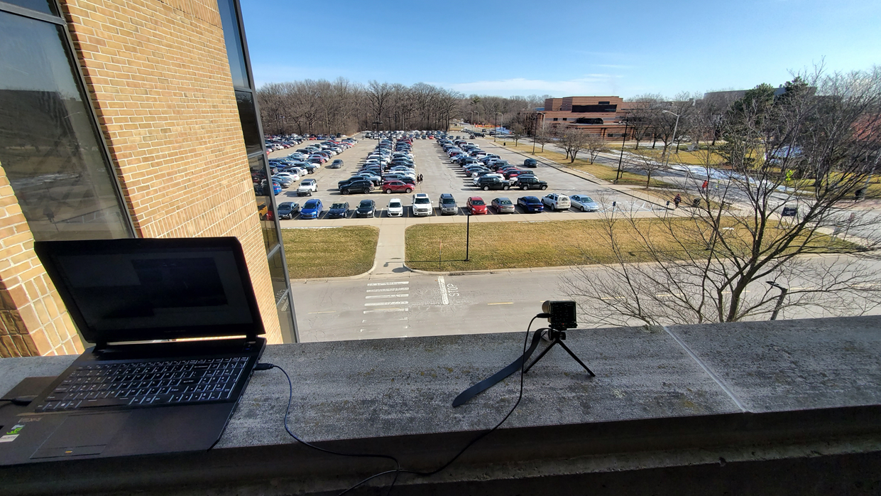}}
    \caption[The data collection setup demonstrating the event camera's placement.]{The data collection setup showing the hybrid event camera (DAVIS 240) mounted on a tripod at the edge of a building overlooking the street and part of the parking lot. A laptop adjacent to the camera setup is used for data recording and sensor control.}
    \label{fig:Dataset: data_collection_setup}
\end{figure}

During the data collection process, the event camera was placed on the edge of a building while pointing downward at the street, representing an infrastructure or traffic surveillance camera setting, as demonstrated in Figure \ref{fig:Dataset: data_collection_setup}. The camera is fixed and kept static throughout (\ie, no ego motion is applied to the camera). Accordingly, the events captured would be only due to an object’s motion or due to noise. Additionally, the standard lens, shipped with the sensor, is manually tuned to enable viewing angles and fields of view as shown in Figure \ref{fig:Dataset:scenes}.

\begin{figure}[t]
    \centering
    \begin{subfigure}[b]{0.32\textwidth}
         \centering
         \frame{\includegraphics[width=\textwidth]{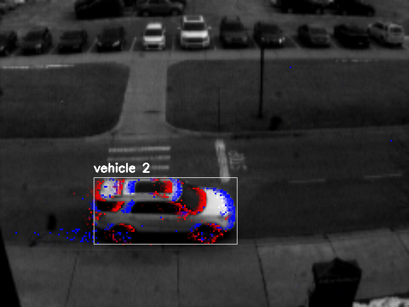}}
         \caption{}
    \end{subfigure}
    \begin{subfigure}[b]{0.32\textwidth}
         \centering
         \frame{\includegraphics[width=\textwidth]{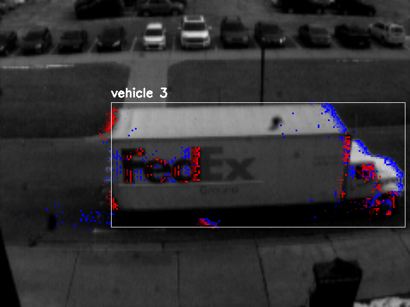}}
         \caption{}
    \end{subfigure}
    \begin{subfigure}[b]{0.32\textwidth}
         \centering
         \frame{\includegraphics[width=\textwidth]{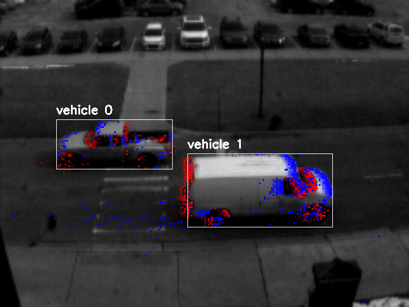}}
         \caption{}
    \end{subfigure}
    \smallskip
    
    \begin{subfigure}[b]{0.32\textwidth}
         \centering
         \frame{\includegraphics[width=\textwidth]{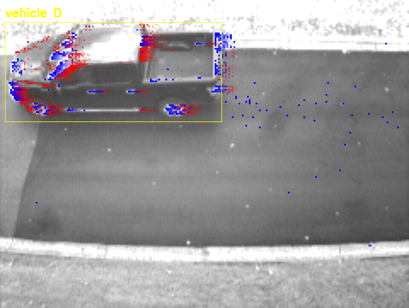}}
         \caption{}
    \end{subfigure}
    \begin{subfigure}[b]{0.32\textwidth}
         \centering
         \frame{\includegraphics[width=\textwidth]{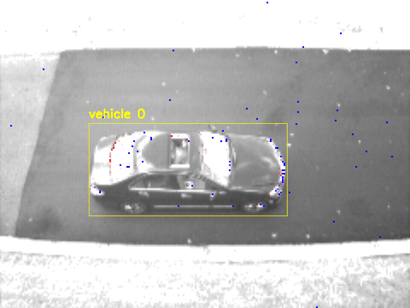}}
         \caption{}
    \end{subfigure}
    \begin{subfigure}[b]{0.32\textwidth}
         \centering
         \frame{\includegraphics[width=\textwidth]{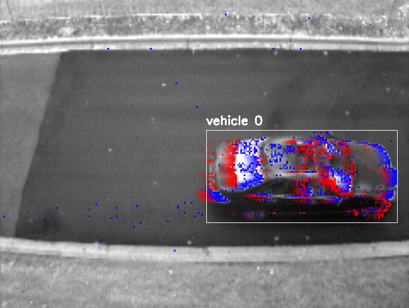}}
         \caption{}
    \end{subfigure}    

    \caption{Samples from the dataset showing labeled vehicles. Each image demonstrates the \ac{APS} intensity frame with superimposed events from the \ac{DVS} collected in the last $\sim$43 ms, where blue and red pixels visualize positive and negative events, respectively. The samples include various vehicle types such as (a) SUVs, (b) trucks, (c) vans, and (d) pickup trucks captured in two different scenes (Scene A for the top row and Scene B for the bottom row). The presence of multiple objects and vehicles at different speeds (e--f) illustrates the dataset's utility for object detection and tracking research.}
    \label{fig:Dataset:labeled_samples}
\end{figure}

In this dataset, we focused on capturing sequences of moving vehicles of different types (\eg, sedans, trucks, etc.), as shown in Figure \ref{fig:Dataset:labeled_samples}. Some recordings of pedestrians passing by were also collected (exclusively available in Scene A's sequences). However, these instances were excluded from the annotation process. Pedestrians were not the focus of this work due to their relatively slow movements and their far proximity to the camera, making the \ac{CV} task of object detection challenging and intermittent. We also note that the vehicles that passed by in the scene did so at varying speeds and accelerations, some reaching a full stop at several instances, thus making the tasks of object detection and tracking more challenging when only employing the event data (captured by the \ac{DVS}).

\subsection{Data Processing and Labeling}\label{sec:data_labeling}



Labeling was manually performed for each vehicle within the scene using the online \textit{dLabel Annotation Tool}\footnote{Available at \href{https://dlabel.org/}{https://dlabel.org/}}. This tool was selected for its precision and ease of use in annotating objects for object detection and tracking applications. It supports sub-pixel accuracy in annotations and includes features such as bounding box interpolation across sequences, which is quite useful for annotating unique objects that appear in consecutive images, thereby significantly reducing the effort and time required for manual labeling. Each vehicle instance was carefully marked with a \ac{2D} bounding box and assigned a unique object ID to ensure tracking continuity using the intensity images captured by the \ac{APS}. This process resulted in a labeling frequency of $\sim24$ Hz, matching the framerate of the camera's \ac{APS}.

After the annotation process, the resulting labels were initially saved in the COCO format \cite{lin2014microsoft} and subsequently converted to the various formats and temporal resolutions detailed in Section \ref{sec:label_format} using custom scripts developed specifically for this dataset. These scripts utilize the data's microsecond timestamps to interpolate the ground truth labels, thus generating high-temporal-resolution labels that significantly enhance the dataset’s utility for high-precision object tracking.



Labels are directly transferable to the event-based modality thanks to the temporal and spatial synchronization between the \ac{APS} and \ac{DVS}. The temporal synchronization is enabled by \ac{DAVIS}'s high-resolution clock \cite{DAVIS240brandli2014240}, while the spatial synchronization is facilitated by the shared lens between the camera's \ac{APS} and \ac{DVS}. This synchronization ensures that a pixel $(x_i, y_i)$ in one modality precisely corresponds to the same pixel in the other, significantly enhancing the dataset’s utility for cross-modal and multi-modal studies by allowing annotations to be used seamlessly across both.

The labeled \ac{MOT} data provides true \ac{2D} bounding boxes for all vehicles in the scene present in any image, along with their corresponding object IDs, essential for proper object tracking evaluation. In contrast, the object detection labels, as detailed in Section \ref{sec:label_format}, include the object classification index and \ac{2D} bounding box coordinates.
Figure \ref{fig:Dataset:labeled_samples} demonstrates several samples from our dataset, showcasing ground truth annotations with objects’ bounding boxes and unique IDs, with the latest $\sim$43 ms of events superimposed on each grayscale image.


In the \ac{MEVDT} dataset, our labeling efforts primarily focused on moving vehicles. Parked vehicles in Scene A, shown in Figure \ref{fig:Scene_A}, were intentionally excluded to maintain the dataset's emphasis on dynamic scenarios. These vehicles, at the top of the frame, were not labeled due to their static nature and relative size, including any moving vehicles behind them. Users are advised to crop or ignore the upper 15--20\% of the frame in Scene A during training or fine-tuning deep learning-based models to avoid these static vehicles. Alternatively, the detections of the parked vehicles can be disregarded when using off-the-shelf pre-trained object detectors, such as YOLOv3 \cite{redmon2018yolov3}, as implemented in our prior research \cite{el2022high, el2023high}.


While our labeling efforts concentrated on moving vehicles, some sequences in Scene A do include pedestrians. These pedestrians were not labeled due to their relatively low number and infrequent appearance, aligning with our dataset's focus on more dynamic and prevalent traffic elements for object tracking and detection tasks. Future data collection efforts should consider locations with a higher number of active and mobile pedestrians.

\section*{Limitations}

Although the \ac{MEVDT} dataset provides valuable data for event-based vision research, it has certain limitations. The dataset focuses exclusively on vehicles, with no labeled pedestrians, which reduces object variety and may affect the generalizability of models. Due to the proximity of the camera, the object sizes are generally uniform with relatively similar viewing perspectives, though some variations exist between scenes. Additionally, the camera remains fixed throughout the data collection process, resulting in negligible ego-motion --- an essential aspect for some applications, such as \ac{AD}. The dataset's scale, though substantial, may be insufficient for training highly complex models or tasks requiring large-scale data. Finally, environmental variations, such as lighting and weather conditions, are limited, potentially impacting the robustness of models in diverse real-world scenarios.

\section*{Ethics Statement}

The authors confirm they have read and followed the ethical requirements for publication in Data in Brief and confirm that the current work does not involve human subjects, animal experiments, or any data collected from social media platforms.

\section*{CRediT Author Statement}



\textbf{Zaid A. El Shair:} Conceptualization, Data curation, Investigation, Methodology, Software, Visualization, Writing; \textbf{Samir A. Rawashdeh:} Conceptualization, Methodology, Project administration, Resources, Supervision, Validation.

\section*{Data Availability}
\href{https://doi.org/10.7302/d5k3-9150}{MEVDT: Multi-Modal Event-Based Vehicle Detection and Tracking Dataset} (Deep Blue Data).

\section*{Acknowledgements}

We sincerely thank Mariana Al Bader for her efforts in carefully annotating the majority of this dataset's samples.

This research did not receive any specific grant from funding agencies in the public, commercial, or not-for-profit sectors.

\section*{Declaration of Competing Interests}
The authors declare that they have no known competing financial interests or personal relationships that could have appeared to influence the work reported in this paper.

\bibliographystyle{elsarticle-num} 
\bibliography{references}

\end{document}